\documentclass[10pt,twocolumn,letterpaper]{article}
\usepackage{cuted}
\usepackage{cvpr}              
\usepackage{CJKutf8}
\usepackage{float}

\definecolor{cvprblue}{rgb}{0.21,0.49,0.74}
\usepackage[pagebackref,breaklinks,colorlinks,citecolor=cvprblue]{hyperref}
\usepackage{amssymb}
\usepackage{amsmath}
\usepackage{booktabs}
\usepackage{graphicx}
\usepackage{tabularx} 
\usepackage{multirow}
\usepackage{xcolor}

\def\paperID{1898} 
\def\confName{CVPR}
\def\confYear{2025}

\title{PhotoDoodle: Learning Artistic Image Editing from Few-Shot Examples}

\author{
Shijie Huang\textsuperscript{1,5}\thanks{Equal contribution.} \quad Yiren Song\textsuperscript{1,5}\footnotemark[1] \quad Yuxuan Zhang\textsuperscript{2,5} \quad Hailong Guo\textsuperscript{3,5} \\
Xueyin Wang\textsuperscript{4} \quad Mike Zheng Shou\textsuperscript{1}\thanks{Corresponding author.} \quad Jiaming Liu\textsuperscript{5}\thanks{Project leader.} \\
\textsuperscript{1}National University of Singapore \quad \textsuperscript{2}Shanghai Jiao Tong University \\
\textsuperscript{3}Beijing University of Posts and Telecommunications \quad \textsuperscript{4}Byte Dance \\ \textsuperscript{5}Tiamat \\
}

\begin{document}
\maketitle

\begin{strip}
\centering
    \vspace{-1.5cm}
    \includegraphics[width=\linewidth]{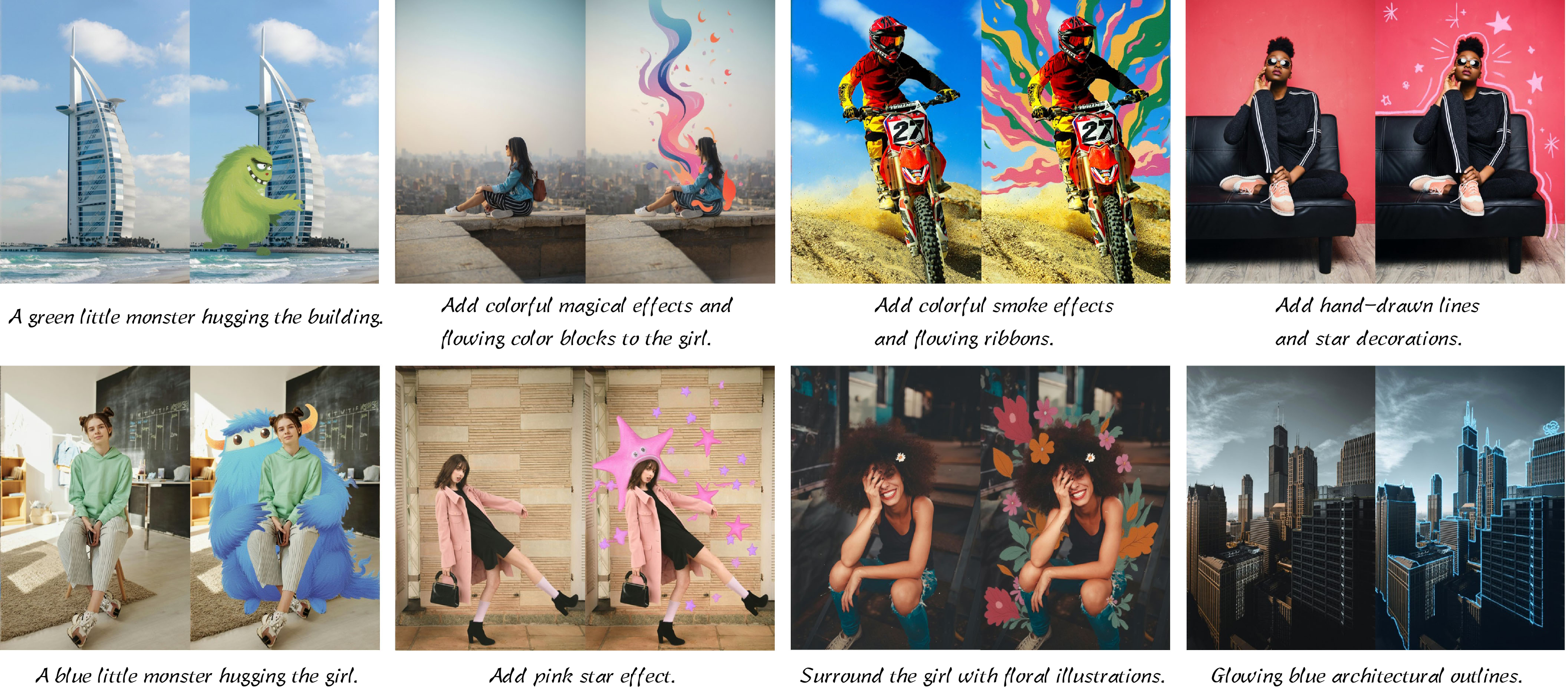}
    \vspace{-7pt}
    \captionof{figure}{PhotoDoodle can mimic the styles and techniques of human artists in creating photo doodles, adding decorative elements to photos while maintaining perfect consistency between the pre- and post-edit states.}
    \label{fig:teaser}
\end{strip}

\begin{abstract}


We introduce PhotoDoodle, a novel image editing framework designed to facilitate photo doodling by enabling artists to overlay decorative elements onto photographs. Photo doodling is challenging because the inserted elements must appear seamlessly integrated with the background, requiring realistic blending, perspective alignment, and contextual coherence. Additionally, the background must be preserved without distortion, and the artist’s unique style must be captured efficiently from limited training data. These requirements are not addressed by previous methods that primarily focus on global style transfer or regional inpainting. The proposed method, PhotoDoodle, employs a two-stage training strategy. Initially, we train a general-purpose image editing model, OmniEditor, using large-scale data. Subsequently, we fine-tune this model with EditLoRA using a small, artist-curated dataset of before-and-after image pairs to capture distinct editing styles and techniques. To enhance consistency in the generated results, we introduce a positional encoding reuse mechanism. Additionally, we release a PhotoDoodle dataset featuring six high-quality styles. Extensive experiments demonstrate the advanced performance and robustness of our method in customized image editing, opening new possibilities for artistic creation. Code is released at \href{https://github.com/showlab/PhotoDoodle}{https://github.com/showlab/PhotoDoodle}.

\end{abstract}

\section{Introduction}

The rise of diffusion models has started a new chapter for image creation and control. Using the pretrain-finetune approach, the community has achieved remarkable progress in customized image generation \cite{ruiz2023dreambooth, TI, lora}, with applications spanning identity preservation \cite{facechain}, artistic stylization \cite{song2024processpainter, zhang2023inversion,sohn2023styledrop,ahn2024dreamstyler}, and subject coherence \cite{customdiff,ruiz2023dreambooth, TI,lora,ruiz2024hyperdreambooth,jiang2024videobooth,zhu2024multibooth}. These advancements have fueled applications ranging from digital art creation to commercial design.

Despite these successes, there remains a critical gap between customized image generation and image editing. Current methods primarily focus on content creation, while intelligent context-aware editing, particularly for artistic enhancement, remains underexplored. This imbalance stands in contrast to the growing demand for precision image editing tools, with current approaches focusing on content generation rather than context-aware editing. As a result, this underexplored frontier of customized image editing is both a technical challenge and a key to unlocking transformative applications.

PhotoDoodling, as a paradigmatic challenge in customized image editing, demands artists to enhance background photographs through strategic integration of decorative elements (e.g., stylized linework, ornamental patterns) and context-aware modifications for personalized aesthetics. Conventional workflows involve artistic techniques, such as: (1) local stylization, (2) decorative contour rendering, (3) semantic-aware object insertion, and (4) ornamental augmentation. While these processes showcase distinctive artistic signatures and strategic design logic, their manual execution incurs prohibitive time costs, fundamentally constraining both production scalability and the curation of large-scale paired training datasets required for data-driven approaches. However, automating these workflows introduces three interlocked technical barriers: First, harmonious integration demands generated decorations to simultaneously satisfy perspective alignment and semantic coherence with background contexts. Second, strict background preservation requires mechanisms to prevent unintended changes, such as color distribution shifting and texture pattern alternation. Third, efficient style distillation must extract artists' unique editing patterns from sparse pairwise examples (30-50 image pairs). These compounded challenges cause existing methods to be incompetent in addressing the problem in a comprehensive way.

Prevailing image editing paradigms can hardly deal with these challenges altogether. Global editing methods (e.g., Prompt-to-Prompt\cite{p2p}, InstructP2P\cite{brooks2023instructpix2pix}), while effective for consistent style transfer, inadvertently distort background content during localized modifications. Inpainting-based approaches\cite{zhang2024magicbrush,stylizedneuralpainting}, though capable of preserving unmasked regions through localized editing, impose impractical demands for pixel-perfect user-defined masks, fundamentally conflicting with the need for automatic PhotoDoodling. To bridge this gap, we introduce an instruction-guided image editing framework that discarded mask dependency, enabling precise and style-conscious decorative generation while ensuring background consistency.

Our proposed framework, PhotoDoodle, presents a few-shot artistic image editing framwork built upon Diffusion Transformers (DiT), featuring a dual-stage training architecture. In the first phase, we evolve a pre-trained text-to-image DiT model into a universal image editor (OmniEditor) through two key innovations: (1) a Positional Encoding (PE) Cloning mechanism that preserves spatial fidelity by providing coordinate-aware hints, and (2) a noise-free conditioning paradigm that offers non-distorted information of the source image. This foundational stage is trained on 3.5M image editing pairs\cite{ge2024seed}, establishing robust general editing capabilities. The second phase introduces an EditLoRA module that distills artist-specific editing patterns from merely 30-50 exemplar pairs through low-rank adaptation(LoRA), enabling efficient style customization while maintaining base model's capability. This co-designed architecture ensures a balance between artistic flexibility and strict consistency.




In summary, our contributions are threefold:
\begin{itemize}
    \item First framework for artistic photo-doodling: A DiT-based architecture to enable few-shot learning of style-specific editing operations while preserving background integrity.
    \item We propose a noise-free conditioning paradigm with positional encoding cloning for implicit feature alignment, enabling high-fidelity image editing through EditLoRA-enhanced Diffusion Transformers (DiTs) that efficiently learn customized operations while maintaining strict background consistency
  
    \item We collected the first publicly available curated photo-doodle dataset comprising 300+ high-quality pairs across 6 artistic styles, establishing a benchmark for reproducible research.

\end{itemize}

\section{Related work}
\subsection{Diffusion Model and Conditional Generation}

Recent advances in diffusion models have significantly advanced the state of text-conditioned image synthesis, achieving remarkable equilibrium between generation diversity and visual fidelity. Pioneering works such as GLIDE \cite{nichol2021glide}, hierarchical text-to-image models \cite{ramesh2022hierarchical}, and photorealistic synthesis frameworks \cite{saharia2022photorealistic} have systematically addressed key challenges in cross-modal generation tasks. The emergence of Stable Diffusion \cite{rombach2022high}, which implements a Latent Diffusion Model with text-conditioned UNet architecture, has established new benchmarks in text-to-image generation and inspired subsequent innovations including SDXL \cite{podell2023sdxl}, GLiGEN \cite{li2023gligen}, and Ranni \cite{feng2023ranni}. To enhance domain-specific adaptation, parameter-efficient fine-tuning techniques like Low-Rank Adaptation (LoRA) \cite{lora} and DreamBooth \cite{ruiz2023dreambooth} have demonstrated effective customization of pre-trained models. Concurrently, research efforts have focused on precise control over pictorial concepts through multi-concept customization \cite{kumari2023multi}, image prompt integration \cite{ye2023ip，ssr, fast_icassp}, and identity-preserving generation \cite{wang2024instantid}. The introduction of ControlNet \cite{zhang2023adding} further extended controllable generation capabilities to spatial constraints and depth information, with subsequent extended applications on various scenarios\cite{zhang2024stablehair, wang2024stablegarment,zhang2024stablemakeup,guo2025any2anytryon}. Some work \cite{antidreambooth, ringid, antirefernece,  idprotector} also focuses on the security issues of customized generation. Building upon these foundations, our work investigates the novel application of pre-trained diffusion transformers for generating artistic photo-doodles. Unlike previous approaches focusing on photorealism or explicit control modalities, we explore the model's potential in capturing freehand artistic expression while maintaining structural coherence.


\begin{figure*}[ht]
    \centering
    \includegraphics[width=1.0\linewidth]{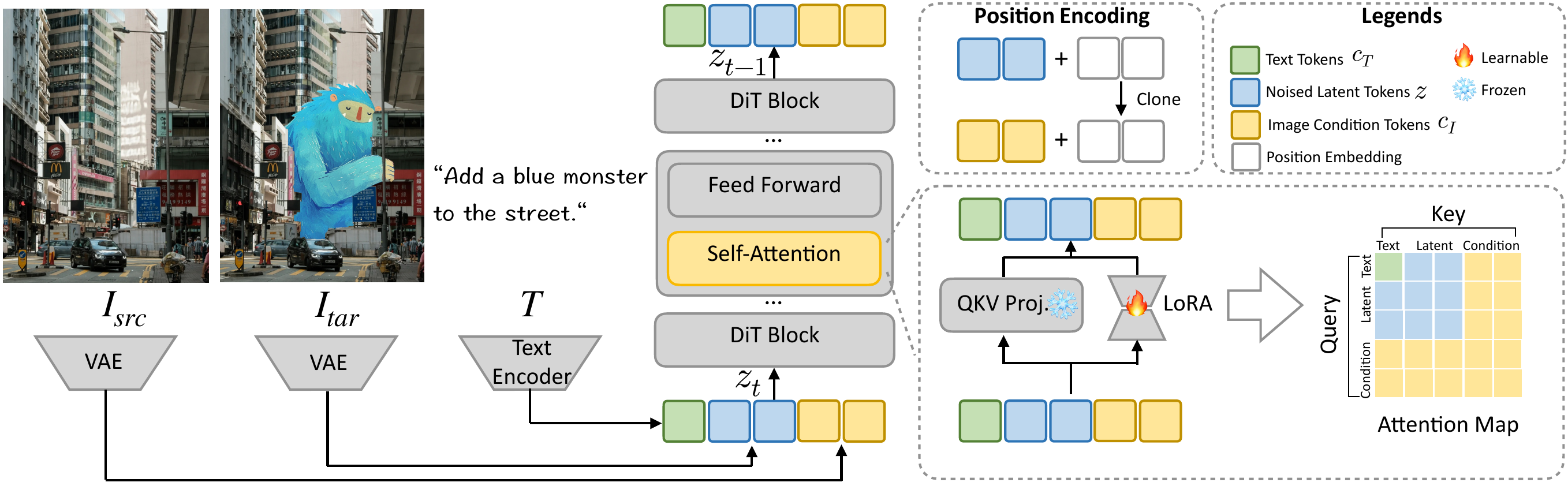} 
    \caption{The overall architecture and training prodigim of photodoodle. The ominiEditor and EditLora all follow the lora training prodigm. We use a high rank lora for pre-training the OmniEditor on a large-scale dataset for general-purpose editing and text-following capabilities, and a low rank lora for fine-tuning EditLoRA on a small set of paired stylized images to capture individual artists’ specific styles and strategies for efficient customization. We encode the source image into a condition token and concatenate it with a noised latent token, controlling the generation outcome through MMAttention.}
    \label{fig2}
\end{figure*}

\subsection{Text-guilded Image Editing}



Recent advances in text-guided image editing have established this field as a critical research frontier in visual content manipulation, with current methodologies generally classified into three paradigms: global description-guided, local description-guided, and instruction-guided editing. Global description-guided approaches (e.g., Prompt2Prompt \cite{p2p}, Imagic \cite{kawar2023imagic}, EditWorld \cite{editworld}, ZONE \cite{li2024zone} ), achieve fine-grained manipulation through cross-modal alignment between textual descriptions and image regions, yet demand meticulous text specification for target attributes. Local description-guided methods such as Blended Diffusion \cite{avrahami2022blended} and Imagen Editor \cite{imageneditor} enable pixel-level control via explicit mask annotations and regional text prompts, though their practical application is constrained by the requirement for precise spatial specifications, particularly in complex editing scenarios like object removal. The emerging instruction-guided paradigm, exemplified by InstructPix2Pix \cite{brooks2023instructpix2pix} and HIVE \cite{zhang2024hive}, represents a paradigm shift through its natural language interface that accepts editing commands (e.g., "change the scene to spring"). This approach eliminates the dependency on detailed textual descriptions or precise mask annotations, significantly enhancing user accessibility.

\subsection{Image and video Doodles}


Image and video doodling involve the creative process of adding hand-drawn elements or animations to static images or video content, blending aspects of graphic design, illustration, and animation to produce playful and engaging visual styles. In academic research, advanced techniques\cite{yu2023designing,videodoodles,belova2021google} have been developed to automate parts of this process. These methods enable users to generate intricate doodle animations from simple textual descriptions, sketches, or keyframes. By preserving artistic intent and reducing the time and expertise required, these models make multimedia content creation more accessible and efficient. Video doodling extends these capabilities to dynamic video sequences, allowing static doodle elements to be seamlessly integrated and animated in response to video motion and context. This innovation not only enhances interactivity but also introduces greater complexity and realism, making it a powerful tool for creative applications in entertainment, education, and storytelling.

\section{Method}
In this section, we begin by exploring the preliminaries on diffusion transformer as detailed in Section 3.1. Next, Section 3.2 outlines our three-stage system design. We then detail two key innovations—OmniEditor Pre-training (Section 3.3) and EditLoRA for style adaptation (Section 3.4). Finally, Section 3.5 describes how we built the PhotoDoodle dataset.

\subsection{Preliminary}

The Diffusion Transformer (DiT)\cite{dit} powers modern image generators like Stable Diffusion 3\cite{sd3} and PixArt\cite{chen2023pixartalpha}. At its core, DiT uses a special transformer network that denoise the noisy image tokens step-by-step.

DiT operates on two categories of tokens: noisy image tokens $z \in \mathbb{R}^{N \times d}$ and text condition tokens $c_T \in \mathbb{R}^{M \times d}$, where $d$ denotes the embedding dimension, and $N$ and $M$ represent the numbers of image and text tokens, respectively. These tokens retain their shapes consistently as they propagate through multiple transformer layers.

In FLUX.1, each DiT block incorporates layer normalization, followed by Multi-Modal Attention (MMA) \cite{sd3}, which employs Rotary Position Embedding (RoPE) \cite{rope} to encode spatial information. For image tokens $z$, RoPE applies rotation matrices based on the token’s position $(i,j)$ in the 2D grid:
\begin{equation}
z_{i,j} \rightarrow z_{i,j} \cdot R(i,j),
\end{equation}
where $R(i,j)$ represents the rotation matrix corresponding to position $(i,j)$. Similarly, text tokens $c_T$ are transformed with their positions designated as $(0,0)$.

The multi-modal attention mechanism projects these position-encoded tokens into query $Q$, key $K$, and value $V$ representations, enabling attention computation across all tokens:
\begin{equation}
\text{MMA}([z; c_T]) = \text{softmax}\left(\frac{QK^\top}{\sqrt{d}}\right)V,
\end{equation}
where $\mathit{Z}=[z; c_T]$ denotes the concatenation of image and text tokens. This approach ensures bidirectional attention between the tokens.


\subsection{Overall Architecture}

The overall architecture of PhotoDoodle is illustrated in Fig. \ref{fig2}. It comprises the following stages:

\noindent \textbf{Pre-training OmniEditor.} The OmniEditor is pre-trained on a large-scale image editing dataset to acquire general-purpose image editing capabilities and strong text-following abilities. This stage ensures the model's capability in diverse editing tasks.

\noindent \textbf{Fine-tuning EditLoRA.} After pre-training, EditLoRA is fine-tuned on a small set of pairwise stylized images (20–50 pairs) to learn the specific editing styles and strategies of individual artists. This stage enables efficient customization for personalized editing needs.



\noindent \textbf{Inference.}: During inference, the input source image $I_{src}$ is encoded as condition tokens $c_I$ via VAE. We then randomly sample Gaussian noise as image tokens $z$, cloning the Position Encoding from condition tokens, and concatenate the tokens along the sequence dimension. Subsequently, we apply the flow matching method to predict the target velocity, iterating multiple steps to obtain the predicted image latent representation. Finally, the predicted image tokens are converted by the VAE decoder to achieve the final predicted image. 



\subsection{OmniEditor Pre-training}
We denote the pre-edited image as the source image $I_{src}$ and the post-edited image as the target image $I_{tar}$. Previous works, such as SDEdit, model image editing as an adding-denoising problem, they often altering unintended areas. Others like InstructP2P\cite{brooks2023instructpix2pix} redesign core model parts, significantly degrading the capacity of the pretrained t2i models. Unlike them, our approach, PhotoDoodle, conceptualizes image editing as a conditional generation problem and minimizes the modification of pretrained text-to-image DiT. 

Our model leverages the advanced capabilities of the DiT-based pretrained model, and extends it to function as an image editing tool. Both $I_{src}$ and $I_{tar}$ are encoded into their respective latent representations, $c_I$ and $z$, via a VAE. After applying position encoding cloning, the latent tokens are then concatenated along the sequence dimension to perform joint attention. The Multi-modal attention mechanisms are used to provide conditional information for the denoising of the doodle image. 
\begin{equation}
\text{MMA}([z; c_{i}; c_T]) = \text{softmax}\left(\frac{QK^\top}{\sqrt{d}}\right)V,
\end{equation}
where $\textit{Z}=[z; c_{i}; c_T]$ denotes the concatenation of noised latent tokens, image condition tokens, and text tokens. Here, $c_{I}$ corresponds to $I_{src}$. This formulation enables bidirectional attention, letting the conditional branch and denoise branch interact on demand.



\noindent \textbf{Position Encoding Cloning.}
Existing approaches to conditional image editing often struggle with pixel-level misalignment between input $I_{src}$ and edited output ($I_{tar}$) that undermine visual coherence. To address this fundamental challenge, we propose a novel Position Encoding(PE) Cloning strategy, motivated by the need for implicit spatial correspondence. 

The PE Cloning is a simple yet powerful stragegy, it simply apply the position encoding calculated on $I_{src}$ on both $I_{src}$ and $I_{tar}$. The identical positional encoding serves as a strong hint so that the DiT is capable of learning correct corresponding easily. By enforcing identical positional encodings between the latent representations $c_I$ and $z$, our method establishes a pixel-perfect coordinate mapping that persists throughout the diffusion process. 
This geometric consistency ensures that every edit respects the original image's spatial structure, eliminating ghosting artifacts and misalignments that plague conventional approaches.

\noindent \textbf{Noise-free Conditioning Paradigm.}  
A critical innovation lies in our noise-free conditioning paradigm. We preserve $c_I$ as a reference during the generation of $I_{tar}$. This design choice achieves two objectives through its operational duality.  
First, by maintaining $c_I$ in a noise-free state, we ensure the retention of high-frequency textures and fine structural details from the original image, thereby preventing degradation during iterative denoising. This preservation mechanism acts as a safeguard against the blurring artifacts commonly observed in conventional approaches. Second, the MM attention machanism is flexible enough to choose either to copy from the source or generate new content via instruction, making the model learns to manipulate only designated target regions.  

Through the combined action of position encodings cloning and MMA mechanism, our framework achieves unprecedented precision in localized editing while maintaining global consistency, a balance previously unattainable in conditional image generation tasks.




\noindent \textbf{Conditional flow matching loss.} The conditional flow matching loss function is following SD3 \cite{sd3}, which is defined as follows:
\begin{equation}
L_{CFM} = E_{t, p_t(z|\epsilon), p(\epsilon)} \left[ \left\| v_\Theta(z, t, c_I,c_T) - u_t(z|\epsilon) \right\|^2 \right]
\end{equation}

Where $ v_\Theta(z, t, c_I,c_T)$ represents the velocity field parameterized by the neural network's weights, $t$ is timestep, $c_I$ and $c_T$ are image condition tokens extracted from source image $I_{src}$ and text tokens. $u_t(z|\epsilon)$ is the conditional vector field generated by the model to map the probabilistic path between the noise and true data distributions, and $E$ denotes the expectation, involving integration or summation over time $t$, conditional $z$, and noise $ \epsilon $.

\subsection{EditLoRA }
LoRA \cite{lora} enhances model adaptation by freezing the pre-trained model weights $ W_0 $ and inserting trainable rank decomposition matrices $ A $ and $ B $ into each layer of the model. These matrices, $ A \in \mathbb{R}^{r \times k} $ and $ B \in \mathbb{R}^{d \times r} $, where $ r \ll \min(d, k) $, are used to fit the residual weights adaptively. The forward computation integrates these modifications as follows:
\begin{equation}
y' = y + \Delta y = W_0 x + B A x
\end{equation}
where $ y \in \mathbb{R}^d $ is the output and $ x \in \mathbb{R}^k $ denotes the input. $ B \in \mathbb{R}^{d \times r} $, $ A \in \mathbb{R}^{r \times k} $ with $ r \ll \min(d, k) $. Normally, matrix $ B $ is initialized with zero.

To learn an individual artist's editing style and effectively transfer it from a small number of before-and-after image pairs, we introduce EditLoRA. Inspired by recent low-rank adaptation (LoRA) techniques \cite{makeanything, layertracer}, EditLoRA fine-tunes only a small set of trainable parameters, significantly reducing the risk of overfitting while preserving most of the pretrained model's expressive power. In our work, the general-purpose OmniEditor is trained on a large-scale paired dataset with a higher-rank lora. The EditLoRAs are lower-rank loras that specifically focus on mimicking the style and strategies of single artists in creating photo doodles. EditLoRA's training set consists of before-and-after pairwise data and corresponding text instruction, which differ from the conventional text-image pairs required for learning image generation models.

The EditLora steers the behaviour of the OmniEditor to the specified artist's style. When a new image $I_{src}$ is provided, along with the textual instructions, the model generates $I_{tar}$ that reflects both the previously learned editing capabilities and the distinctive stylistic effects from the artist.





\begin{figure*}[ht]
    \centering
    \includegraphics[width=1.0\linewidth]{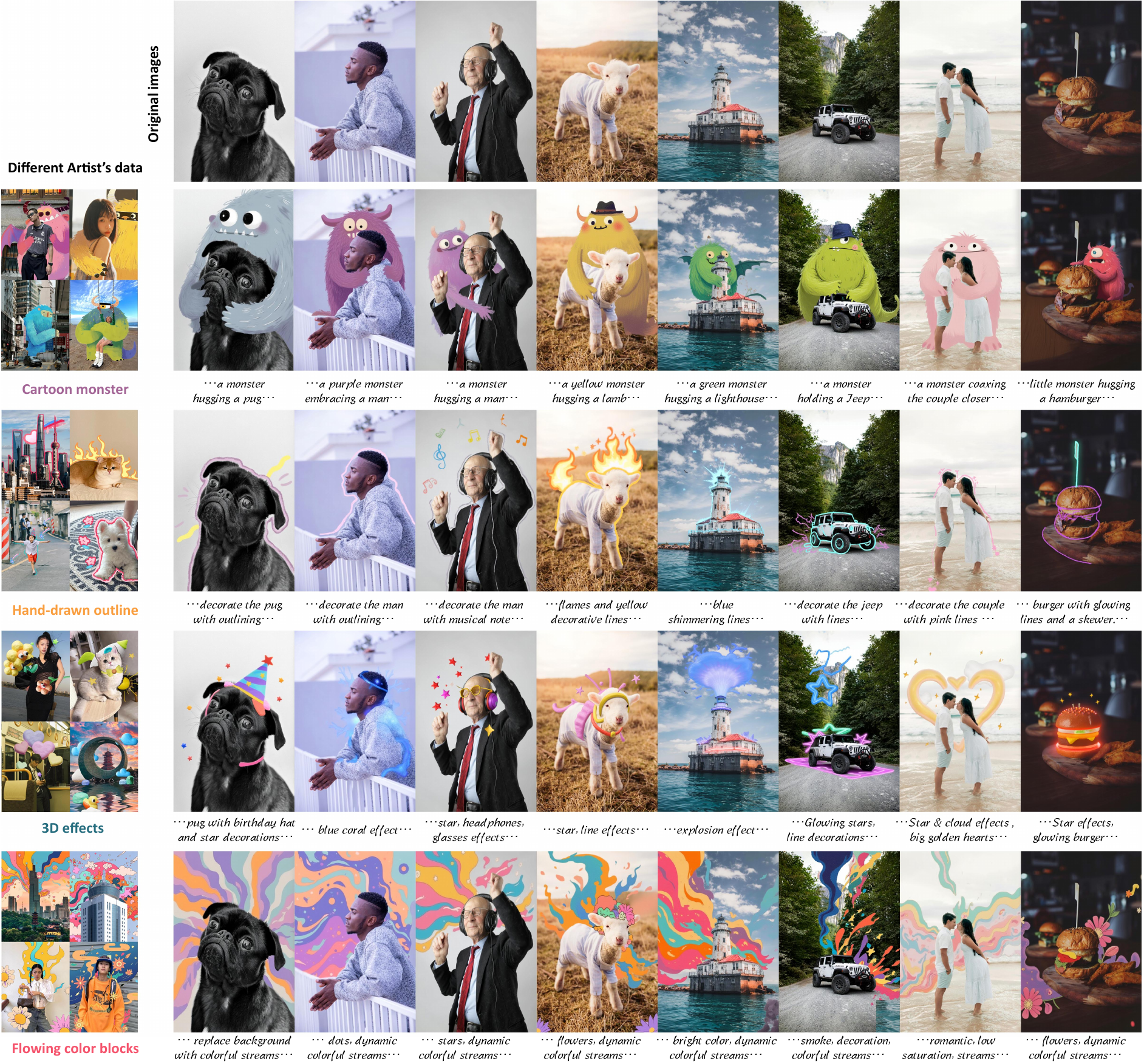} 
    \caption{The generated results of PhotoDoodle. PhotoDoodle can mimic the manner and style of artists creating photo doodles, enabling instruction-driven high-quality image editing.}
    \label{fig3}

\end{figure*}

\begin{figure*}[ht]
    \centering
    \includegraphics[width=1.0\linewidth]{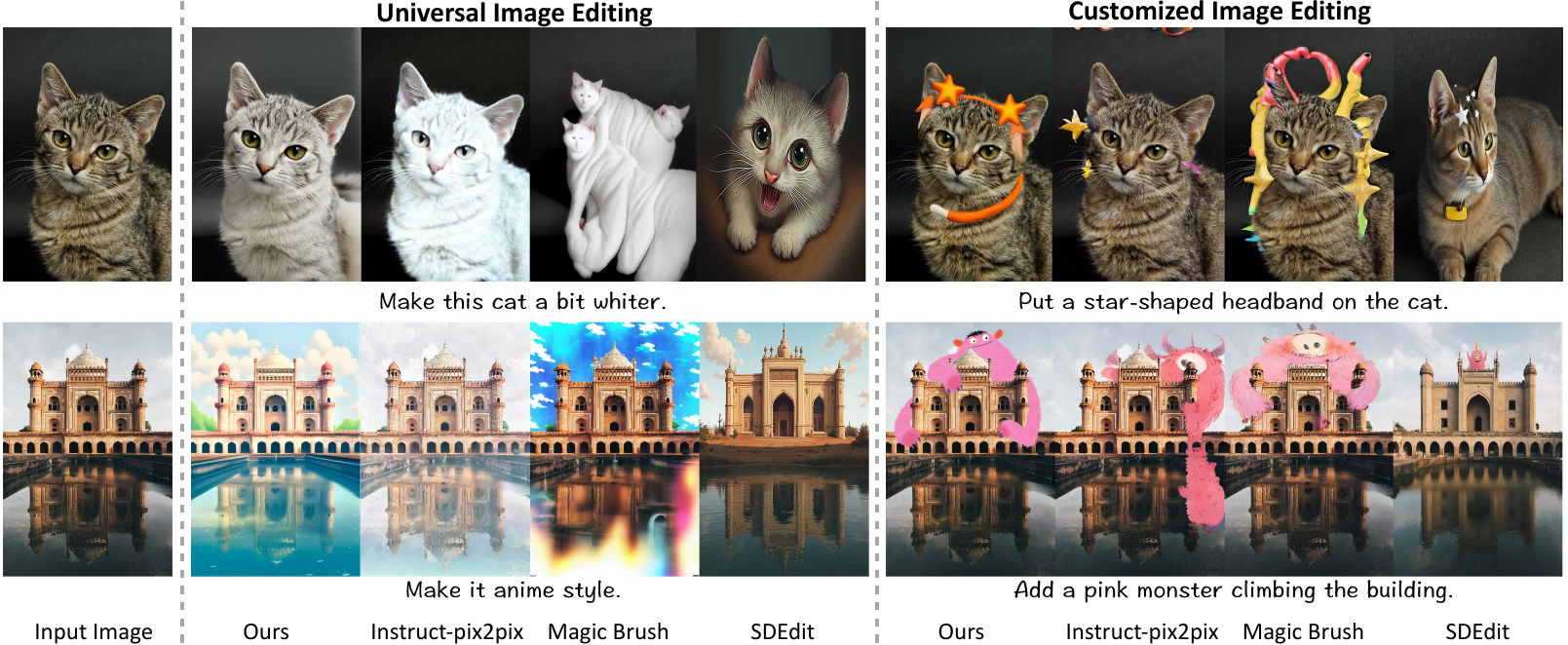} 

    \caption{ Compared to baselines, PhotoDoodle demonstrates superior instruction following, image consistency, and editing effectiveness.}
    \label{fig4}

\end{figure*}

\subsection{PhotoDoodle}

In collaboration with professional artists and designers, we created the first PhotoDoodle dataset. We introduce the dataset containing six high-quality styles and over 300 photo doodle samples. The six styles include cartoon monster, hand-drawn outline, 3D effects, flowing color blocks, flat illustrator, and cloud sketch. Each sample in our dataset consists of a pre-edit photo (e.g., a real-world scene or portrait) and a post-edit photo doodle showing unique modifications by the artist, such as localized stylization, decorative lines, newly added objects, or modifications to existing elements. For each example, we store the raw input image $I_{src}$ and the doodled version $I_{tar}$, along with textual instructions.



\section{Experiment}

\subsection{Experiment Setting}
\noindent \textbf{Setup.}  During the OmniEditor pre-training stage, we take the parameters of Flux.1 dev model as the initialization of the DiT architecture, and trained it with the SeedEdit dataset. Images were resized to 768x512. We trained a LoRA rank of 256, a batch size of 128, and a learning rate of $1 \times 10^{-4}$, on 8 H100 GPUs for 330,000 steps. After merge the lora into the base DiT model, we acquired the OmniEditor model for further usage. In the EditLoRA training phase, we fine-tuned the merged model on a paired photo doodle dataset (50 pairs) using a single GPU for 10,000 steps, with a LoRA rank of 128, batch size of 2, and a learning rate of $1 \times 10^{-4}$.







\noindent \textbf{Baseline Methods.}  The baselines compared in this paper include InstructP2P\cite{brooks2023instructpix2pix}, MagicBrush\cite{zhang2024magicbrush}, and SDEdit\cite{meng2021sdedit} based on Flux. For a fair comparison, tests were conducted in both general image editing and customized editing scenarios. For the general image editing tests, OmniEditor was evaluated against the aforementioned baselines. In the customized editing scenario, we trained Flux LoRA using doodle results created by professional artists and used it alongside SDEdit as a baseline. For InstructP2P and MagicBrush, the attention layers were also fine-tuned with the same doodle dataset. Finally, all the trained LoRA models were compared with the proposed EditLoRA model.

\noindent \textbf{Benchmarks.}  As with previous methods, we tested the performance of the proposed OmniEditor on the HQ-Edit benchmark\cite{hui2024hq}. For the customized generation tasks, this paper introduced a new benchmark collected from the Pexels website, consisting of 50 high-quality photographs of portraits, animals, and architecture.


\subsection{Generation Results}

Fig. \ref{fig3} and Fig. \ref{fig7} displays the image editing results of PhotoDoodle, which excels in text following due to training on a large dataset of before-after pairs. The generated doodles blend well with the original images. When trained on a limited dataset of artist-paired data using EditLoRA, PhotoDoodle consistently produces artistic doodles while preserving image consistency and avoiding unwanted changes. Notably, our method maintains stable performance and a high success rate, making it suitable for production use and reducing the need for selective sampling.


\begin{table}[ht]
\centering
\footnotesize 
\caption{Comparison Results in General Image Editing Tasks. The best results are denoted as \textbf{Bold}.}

\begin{tabular}{lp{2.0cm}p{1.5cm}p{1.5cm}p{1.5cm}}
\toprule
& \textbf{Methods}   & \textbf{$CLIP\ Score$}↑ & \textbf{$GPT\ Score$}↑  & \textbf{$CLIP_{img}$}↑ \\ \midrule
&Instruct-Pix2Pix    & 0.237    &  38.201                          & 0.806                     \\ 
&Magic Brush         & 0.234    & 36.555                           & 0.811                     \\ 
&SDEdit(FLUX)        & 0.230     & 34.329                              & 0.704                     \\ 
& Ours               & \textbf{0.261}    & \textbf{51.159}                           & \textbf{0.871}                     \\ \midrule
\end{tabular}
\label{tab:comparison_universal}
\end{table}

\subsubsection{Qualitative Evaluation}

In this section, we present the results of the qualitative analysis. As illustrated in Fig.~\ref{fig4}, OmniEdito demonstrates superior consistency and minimizes unintended alterations in general image editing tasks compared to state-of-the-art (SOTA) methods. This performance is attributed to the use of high-quality datasets and a thoughtfully designed model architecture. For custom image editing tasks, our method significantly surpasses baseline methods, as evidenced by the high quality of generated outputs and the strong alignment of the editing style with the original artistic intent, while avoiding undesired modifications.




\begin{table}[ht]
\centering
\footnotesize 
\caption{Comparison Results in Customized Image Editing Tasks. The best results are denoted as \textbf{Bold}.}

\begin{tabular}{lp{2.0cm}p{1.5cm}p{1.5cm}p{1.5cm}}
\toprule
& \textbf{Methods}   & \textbf{$CLIP\ Score$}↑ & \textbf{$GPT\ Score$}↑  & \textbf{$CLIP_{img}$}↑ \\ \midrule
&Instruct-Pix2Pix         & 0.249  & 36.359                     & 0.832             \\ 
&Magic Brush              & 0.247     & 38.478               & \textbf{ 0.885}             \\ 
&SDEdit(FLUX)             & 0.209    & 21.793                 & 0.624             \\ 
&Ours                     & \textbf{0.279}   & \textbf{63.207}                & 0.854             \\  \midrule 

\end{tabular}
\label{tab:comparison_customized}
\end{table}

\subsubsection{Quantitative Evaluation}

In this section, we present the quantitative analysis results. Following InstructP2P\cite{brooks2023instructpix2pix}, we compute the $CLIP\ Score$ and $CLIP_{img}$ metrics for both tasks. Furthermore, as proposed in HQ-Edit \cite{hui2024hq}, we employ GPT4-o to evaluate the alignment between text instructions and editing outputs. As shown in Table~\ref{tab:comparison_universal}, our method outperforms all baselines across all metrics in general image editing tasks, achieving the highest $CLIP\ Score$, $GPT\ Score$, and $CLIP_{image}$ Score. In custom image editing tasks, while some models fail to produce meaningful edits, which leads to high $CLIP_{image}$ scores, our method still holds a clear advantage over the baselines. This is evident in the substantial improvements in $GPT Score$ and $CLIP Score$, both of which evaluate the consistency and quality of the generated content in relation to the artist's original work.

\begin{figure}[ht]
    \centering
    \includegraphics[width=1\linewidth]{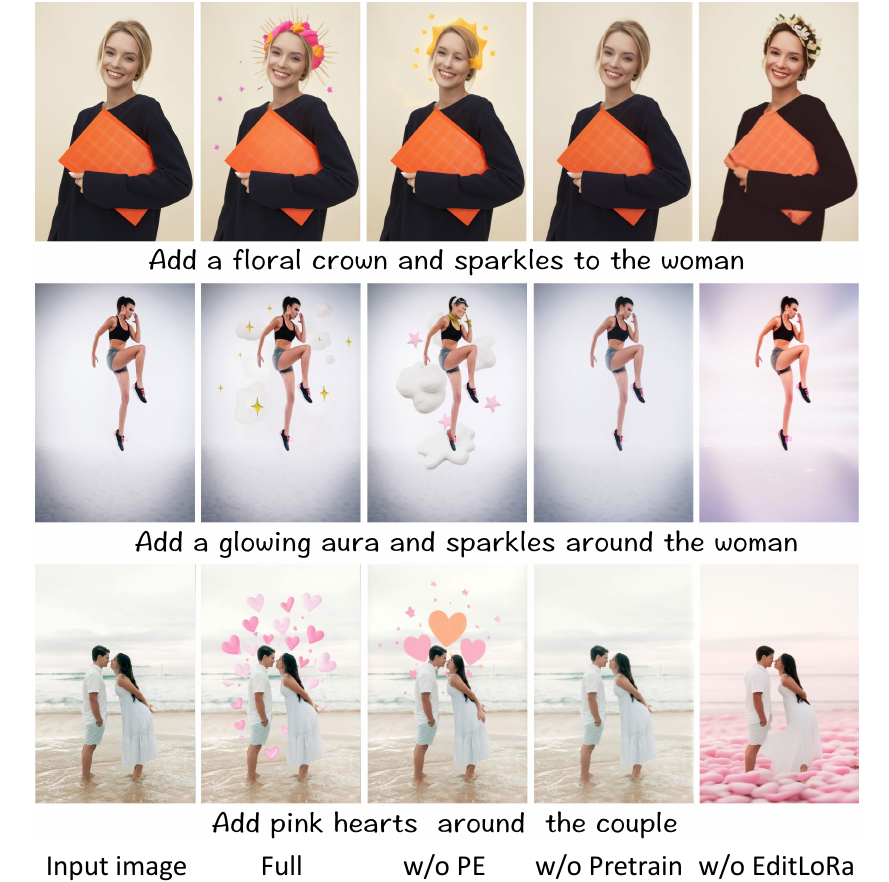} 
    \caption{Ablation study results.}
    \label{fig5}
 \end{figure}


\subsection{Ablation Study}

To demonstrate the effectiveness of the key strategies and modules proposed in this paper, we conducted detailed ablation experiments. We evaluated OmniEditor Pre-training, Position Encoding Cloning, and EditLoRA. As shown in Fig. \ref{fig5}:  Without OmniEditor Pre-training, directly training EditLoRA leads to reduced harmony between the generated sketches and photos, along with weaker text-following capabilities in the output.  Removing Position Encoding Cloning results in decreased consistency in the generated outputs, with unwanted changes occurring in the background.  When EditLoRA is not used, and only the pre-trained OmniEditor is employed for generation, the degree of stylization in the results is significantly reduced.

\subsection{User Study}

\begin{figure}[ht]
    \centering
    \includegraphics[width=1\linewidth]{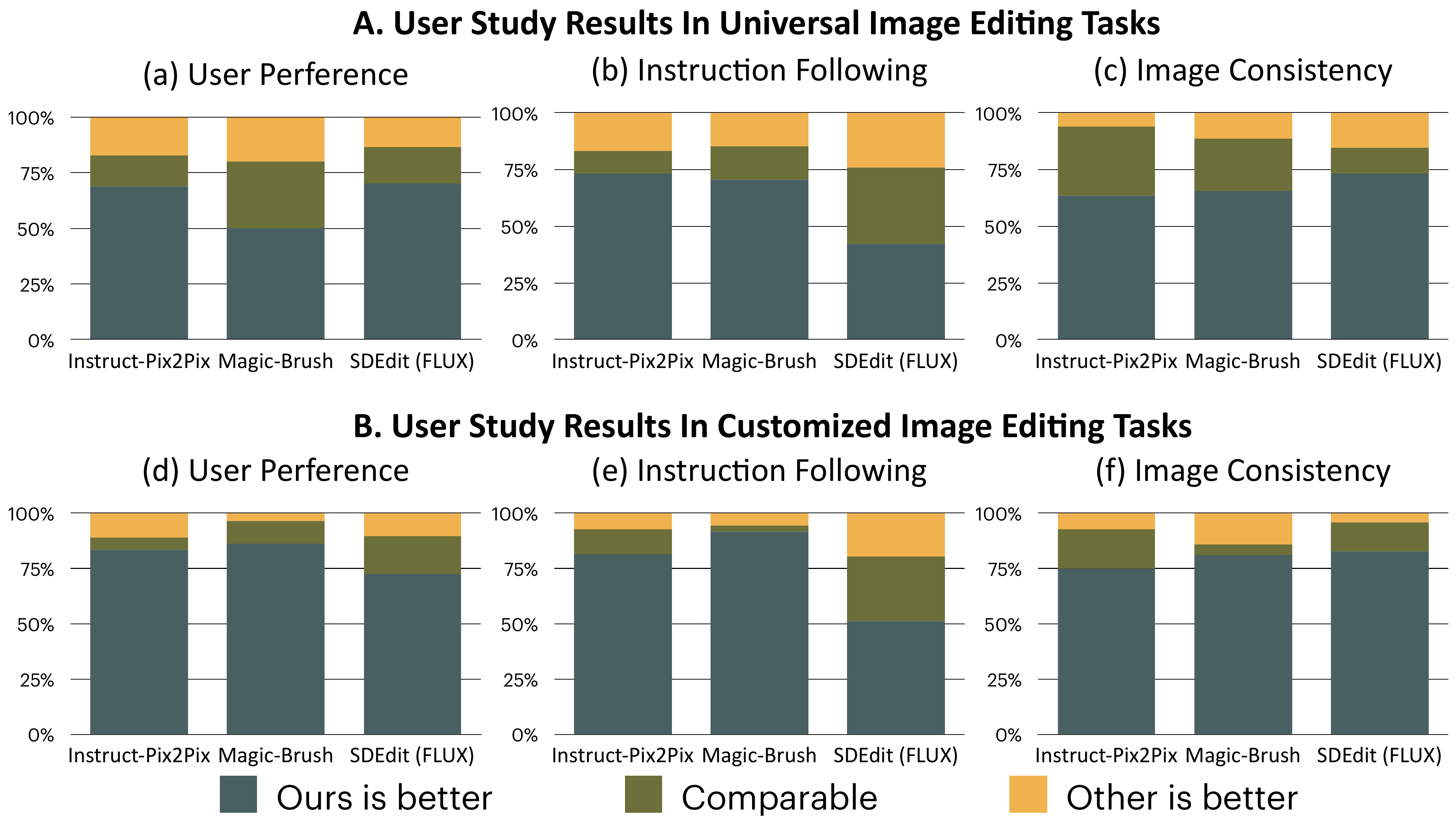} 

    \caption{User study results. The scores demonstrate the percentage of users who prefer ours over others under three evaluation metrics. PhotoDoodle outweighs all other baselines in user study.}

    \label{fig6}
    
\end{figure}

\begin{figure}[ht]
    \centering
    \includegraphics[width=1\linewidth]{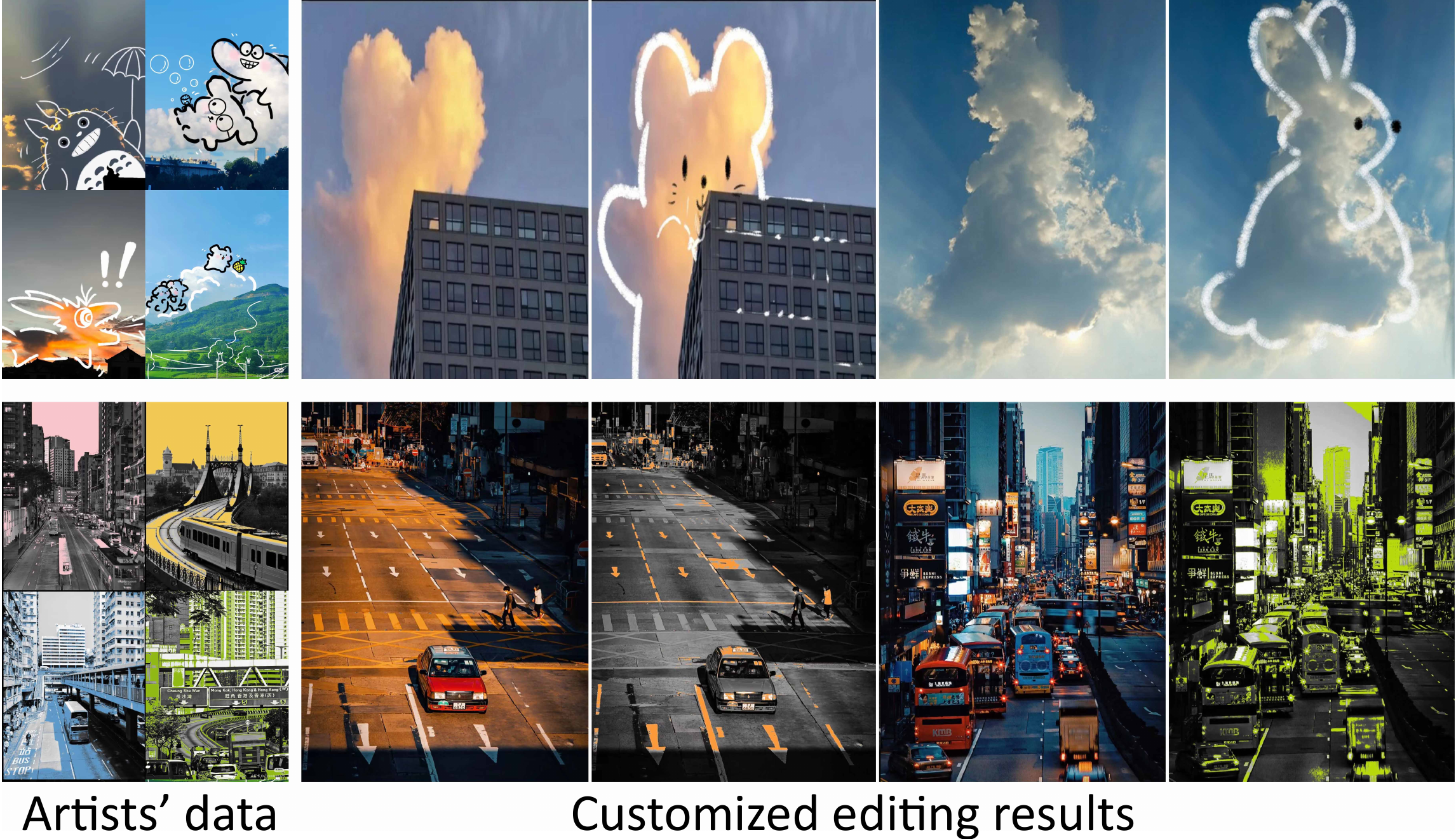} 

    \caption{More photo doodling results: one adds lines to a photo of clouds, imagining them as animals; the other converts the photo into a monochrome version and decorates it with color blocks.}

    \label{fig7}
\end{figure}

To further demonstrate the superiority of our proposed method, we conducted a user study with 30 participants via online questionnaires. We evaluatedd user preferences in both general and customized image editing scenarios. Participants were presented with PhotoDoodle's outputs alongside baseline methods, and asked to evaluate which results they preferred based on three criteria: 1) Overall preference, 2) Instruction following, and 3) Consistency between the edited images and the original images. During the study, participants viewed the original unedited images, the edit instructions, and reference images edited by models. They were then asked to decide whether PhotoDoodle (Option A) or a baseline method (Option B) performed better, or if they were about equally effective. The results of this user study are collected in Fig. \ref{fig6}, where we reported the percentage scores of each criterion, highlighting our method’s effectiveness in aligning closely with artistic intentions and maintaining high consistency in edits without introducing unwanted changes.

\section{Limitation and Future Work}

One limitation of PhotoDoodle is its dependence on the collection of dozens of paired datasets (pre-edit and post-edit images) and the need for thousands of training steps using LoRA. This data collection process can be challenging, as paired images are not always readily accessible. In the future, we will attempt to learn doodling strategies from single image pairs using an Encoder structure.

\section{Conclusion}

In this paper, we present PhotoDoodle, a diffusion-based framework for artistic image editing that learns unique artistic styles from minimal paired examples. By combining large-scale pretraining of the OmniEditor with efficient EditLoRA fine-tuning, PhotoDoodle enables precise decorative generation while maintaining background integrity through positional encoding cloning. Key innovations, including a noise-free conditioning paradigm and parameter-efficient style adaptation requiring only 50 training pairs, significantly reduce computational barriers. We also contribute a new dataset with six artistic styles and 300+ curated samples, establishing a benchmark for reproducible research. Extensive experiments demonstrate superior performance in style replication and background harmony, outperforming existing methods in both generic and customized editing scenarios.

\newpage
\bibliographystyle{ACM-Reference-Format}
\bibliography{main}

{
    \small
    \bibliographystyle{ieeenat_fullname}
    \bibliography{main}
}

\clearpage
\setcounter{page}{1}
\maketitlesupplementary

\section{Experimental Details in Different Bench}
\label{sec:rationale}

On each benchmark, the similarity scores are computed between a reference image and two candidate images, one of which is closer to the reference image. Image pair with the higher score is selected as the choice of current evaluated model. In this section, we will explain details of reference and candidate images selection for each benchmark.

\subsection{NIGHTS Dataset}
NIGHTS (Novel Image Generations with Human-Tested Similarities) is a dataset comprising 20,019 image triplets with human scores of perceptual similarity. Each triplet consists of a reference image and two distortions. This paper utilizes the test set of NIGHTS, which includes 2,120 image triplets. We calculate the DiffSim score for the reference image and the two distortions separately, using human evaluation results as the ground truth.

\subsection{Dreambench++ Dataset}

The Dreambench++ Dataset consists of generated images created using different generation methods, along with human-rated scores for how similar each image is to the original. In our experiment, we use the original image as the reference and randomly select two generated images based on it. The one with the higher human rating is considered closer to the reference. The dataset includes a total of 937 triplets.

\subsection{CUTE Dataset}

The CUTE Dataset includes photos of various instances taken under different lighting and positional conditions. In our experiment, for each category, we repeat the process 10 times: randomly selecting two images of the same instance under the same lighting and one image of a different instance under the same lighting. The two images of the same instance are considered more similar. The dataset contains a total of 1,800 triplets for comparison.

\subsection{IP Bench}

IP Bench contains 299 character classes, each with an original image and six variations generated using different consistency weights. In our experiment, we repeat the process for 5 times: using the original image as the reference and randomly selecting two generated images from the same class. The image with the higher consistency weight is considered closer to the reference. There are a total of 1,495 triplets for comparisons.

\subsection{TID2013 Dataset}

The TID2013 dataset contains 25 reference images, each distorted using 24 types of distortions at 5 different levels. In our experiment, we use a reference image as the starting point and randomly select two distorted images using the same type of distortions from the same reference. The image with a lower distortion level is considered closer to the reference. There are a total of 600 triplets for evaluation.

\subsection{Sref Dataset}

The Sref bench includes 508 styles manually selected by artists and generated by Midjounery, with each style featuring four images. When constructing image triplets, we randomly select two images from the same style and one image from a different style. We fix the random seed to construct 2,000 image triplets for quantitative evaluation.

\subsection{InstantStyle Bench}
The InstantStyle bench includes 30 styles, with each style comprising five images. When constructing image triplets, we randomly select two images from the same style and one image from a different style. We fix the random seed to construct 2,000 image triplets for quantitative evaluation.

\subsection{TikTok Dataset}

For tiktok dataset, we extract 10 frames from each video, and calculate the variance of different similarity metric scores between the first frame and other frame. A lower variance indicates that the metric demonstrates better robustness to changes in the movements of characters in the video.

\begin{table*}[ht]
\centering
\caption{Performance of diffsim across various benchmarks with different pre-trained models. Best results are highlighted in bold.}
\label{tab:5}
\small 
\begin{tabular}{@{}c|cc|cc|c|cc@{}}
\toprule
\textbf{Model / Benchmark} & \multicolumn{2}{c|}{\textbf{Human-align Similarity}} & \multicolumn{2}{c|}{\textbf{Instance Similarity}} & \multicolumn{1}{c|}{\textbf{Low-level Similarity}} & \multicolumn{2}{c}{\textbf{Style Similarity}} \\ 
 & \textbf{NIGHTS} & \textbf{Dreambench++} & \textbf{CUTE} & \textbf{IP} & \textbf{TID2013} & \textbf{Sref}  & \textbf{InstantStyle bench} \\ \midrule
DiffSim-S SD1.5                     & \textbf{86.52\%} & \textbf{71.50}\% & 72.06\% & \textbf{92.04\%} & \textbf{94.17\%} & \textbf{97.40\%} & \textbf{99.05\%} \\
DiffSim-C SD1.5                      & 79.16\% & 67.45\% & \textbf{76.17}\% & 77.06\% & 94.00\% & 94.70\% & 95.10\% \\
DiffSim-S SD-XL                      & 78.05\% & 63.93\% & 69.94\% & 83.41\% & 91.33\% & 93.05\% & 96.55\% \\
DiffSim DIT-XL/2 256                 & 63.38\% & 57.52\% & 53.44\% & 82.81\% & 83.50\% & 77.00\% & 80.15\% \\
DiffSim DIT-XL/2 512                 & 67.92\% & 57.31\% & 57.22\% & 81.00\% & 88.67\% & 78.20\% & 79.40\% \\
\bottomrule
\end{tabular}
\end{table*}

\section{Exploring Different Model Architectures}

In Table ~\ref{tab:5}, we present the performance differences of DiffSim using pre-trained models with different architectures. DiffSim-S SD1.5 leads in all benchmarks except for the CUTE dataset. DiffSim-C SD1.5 performs better on the CUTE dataset, possibly because the cross-attention layers in the U-Net architecture are particularly effective at distinguishing the subject. On the other hand, DiffSim-C uses IP-Adapter Plus, and the CLIP image encoder may become a performance bottleneck in other benchmarks. Models with higher resolution, such as SD-XL and DIT-XL/2 512, do not show performance improvement compared to lower resolution models like SD1.5 and DIT-XL/2 256. Furthermore, the performance of models using DIT as the pre-trained model is worse than using U-Net, with two possible reasons: 1. DIT splits the image into patches and then serializes them, which may lead to the loss of spatial information, which is detrimental to DiffSim, despite the use of positional encoding. 2. DIT is trained on the ImageNet dataset, which is much smaller than the SD1.5 and SD-XL models' training datasets.

\section{Additional Experimental Results}
In Figures \ref{nightsbench} to \ref{instantstylebench} , we present the default implementation of DiffSim, which is based on the self-attention layers of SD1.5, showing results across different layers and denoising time steps t.

\section{Additional Visual Examples}
Figure ~\ref{supp:bench} and ~\ref{supp:IP} show more examples of images from Sref bench and IP bench; Figure ~\ref{supp:retrieval} presents more top-4 retrieval results of DiffSim, CLIP, DINO v2 on MS COCO, Sref bench and IP bench.

\begin{figure}[htp]
    \centering
    \includegraphics[width=1.0\linewidth]{sec/Image/NIGHTS_Dataset_Performance.pdf}
    \caption{Results on NIGHTS dataset.}
    \label{nightsbench}
\end{figure}

\begin{figure}[htp]
    \centering
    \includegraphics[width=1.0\linewidth]{sec/Image/Dreambench_Dataset_Performance.pdf}
    \caption{Results on Dreambench++ dataset.}
    \label{bench}
\end{figure}

\begin{figure}[htp]
    \centering
    \includegraphics[width=1.0\linewidth]{sec/Image/CUTE_Dataset_Performance.pdf}
    \caption{Results on CUTE dataset.}
    \label{bench}
\end{figure}

\begin{figure}[htp]
    \centering
    \includegraphics[width=1.0\linewidth]{sec/Image/IP_Bench_Performance.pdf}
    \caption{Results on IP bench.}
    \label{bench}
\end{figure}

\begin{figure}[htp]
    \centering
    \includegraphics[width=1.0\linewidth]{sec/Image/TID2013_Performance.pdf}
    \caption{Results on TID2013 dataset.}
    \label{bench}
\end{figure}

\begin{figure}[htp]
    \centering
    \includegraphics[width=1.0\linewidth]{sec/Image/Sref_Bench_Performance.pdf}
    \caption{Results on Sref bench.}
    \label{bench}
\end{figure}

\begin{figure}[htp]
    \centering
    \includegraphics[width=1.0\linewidth]{sec/Image/InstantStyle_Bench_Performance.pdf}
    \caption{Results on InstantStyle bench.}
    \label{instantstylebench}
\end{figure}

\begin{figure*}[htp]
    \centering
    \includegraphics[width=0.89\linewidth]{sec/Image/Sref1.jpg}
    \caption{Examples in Sref bench we proposed.}
    \label{supp:bench}
\end{figure*}

\begin{figure*}[htp]
    \centering
    \includegraphics[width=0.89\linewidth]{sec/Image/IPref.pdf}
    \caption{Examples in IP bench we proposed.}
    \label{supp:IP}
\end{figure*}

\begin{figure*}[htp]
    \centering
    \includegraphics[width=0.99\linewidth]{sec/Image/Supp-ImageRetrieval.pdf}
    \caption{More image retrieval results.}
    \label{supp:retrieval}
\end{figure*}


\end{document}